\documentclass[conference]{IEEEtran}
\IEEEoverridecommandlockouts
\usepackage[utf8]{inputenc}

\usepackage{amsmath}

\def\BibTeX{{\rm B\kern-.05em{\sc i\kern-.025em b}\kern-.08em
    T\kern-.1667em\lower.7ex\hbox{E}\kern-.125emX}}
\PassOptionsToPackage{hyphens}{url}\usepackage{hyperref}

\usepackage[T1]{fontenc}

\let\labelindent\relax

\usepackage{hyperref}
\hypersetup{hidelinks}

\usepackage{caption}

\usepackage{graphicx}
\usepackage[caption=false, font=normalsize, labelfont=sf, textfont=sf]{subfig}

\usepackage{tikz} %
\usepackage{epstopdf}
\newcommand{\orcidicon}{\includegraphics[width=0.32cm]{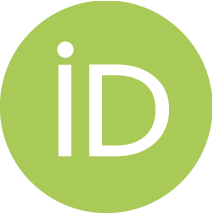}}

\foreach \x in {A, ..., Z}{%
\expandafter\xdef\csname orcid\x\endcsname{\noexpand\href{https://orcid.org/\csname orcidauthor\x\endcsname}{\noexpand\orcidicon}}
}

\usetikzlibrary{decorations.pathreplacing,}

\NewDocumentCommand\mybrace{mmo}{%
\IfValueTF {#3}{%
\begin{tikzpicture}[overlay, remember picture,decoration={brace,amplitude=1ex}]
  \draw[decorate,thick] (#1.north east) -- (#2.south east) node[midway, right=0.1cm] {$=$}node[midway, right=0.5cm,text=black,text width = 2in,] {{#3}};
\end{tikzpicture}%
}%
{%
\begin{tikzpicture}[overlay, remember picture,decoration={brace,amplitude=1ex}]
  \draw[decorate,thick] (#1.north east) -- (#2.south east);
\end{tikzpicture}%
}%
}%

\usepackage{geometry}
\usepackage{etoolbox}
\makeatletter
\patchcmd{\@makecaption}
  {\scshape}
  {}
  {}
  {}
\makeatother

\title{\LARGE \bf
Autonomous Monitoring of Pharmaceutical R\&D Laboratories with 6 Axis Arm Equipped Quadruped Robot and Generative AI: A Preliminary Study}

\author{Shunichi Hato$^{1}$, Nozomi Ogawa$^{2}$
\thanks{$^{1}$Data Sciences Institute, R\&D, Takeda Pharmaceutical Company Limited. $^{2}$Sustainability and Technology, Pharmaceutical Sciences, R\&D, Takeda Pharmaceutical Company Limited, 26-1, Muraoka-Higashi 2-chome, Fujisawa, Kanagawa 251-8555, Japan.
    {\tt\small (\{shunichi.hato, nozomi.ogawa\}@takeda.com})}%
}

\begin{document}

\maketitle
\thispagestyle{empty}
\pagestyle{empty}

\begin{abstract}
This paper presents a proof-of-concept study that examines the utilization of generative AI and mobile robotics for autonomous laboratory monitoring in the pharmaceutical R\&D laboratory. The study investigates the potential advantages of anomaly detection and automated reporting by multi-modal model and Vision Foundation Model (VFM), which have the potential to enhance compliance and safety in laboratory environments. Additionally, the paper discusses the current limitations of the generative AI approach and proposes future directions for its application in lab monitoring.

\end{abstract}


\section{Introduction}  \label{sec:intro} 

A clean and well-organized laboratory is crucial in pharmaceutical research and development as it ensures traceability, minimizes errors and contamination, upholds quality standards, and contributes to regulatory compliance. However, the current reliance on human surveillance for lab monitoring poses challenges in terms of consistent education and supervision.\par
To address these challenges, the integration of computer vision and mobile robotic technology holds promise. By exploring the potential of generative AI with vision capability and mobile robotics, it may be possible to establish a scalable, standardized and routine monitoring system for laboratory environments. Previously, our group reported utilization of quadruped robots in pharmaceutical research and development laboratories for remote inspection using the out-of-box capabilities of Boston Dynamics' Spot platform \cite {Parkinson2023}  \cite {SpotDynamics}. Spot platform is also evaluated similarly for use in inspection and monitoring of construction site \cite {Halder2023ConstructionStudy}. Other mobile platforms have been reported to be utilized for safety inspections in chemistry laboratories, employing infrared thermal imaging and machine vision techniques \cite {mobilerobotIRmachinevision}, as well as for monitoring volatile organic solvents in life science laboratories \cite {mobilerobotVOS}. In this article, we extend our effort to autonomous lab monitoring and present a proof-of-concept study that examines the use of generative AI and mobile robotics. Through the implementation of this technology, laboratories may potentially benefit from real-time monitoring, early anomaly detection, and automated reporting, which could contribute to improved GMP compliance and enhanced safety. 
Specifically, this paper explores the viability of multi-modal models and Vision Foundation Model (VFM) methods for detecting anomalies and levels of organization in lab environments.\par
The evolution of generative AI has ushered in a new era of deep learning, marked by the rise of unsupervised or very few shot methods that obviate the need for extensive training datasets \cite {radford2021learning} \cite {oquab2023dinov2} \cite {he2021masked} \cite {awadalla2023openflamingo} \cite {li2023otter}. Coupled with this, the advent of multi-modal models that can process and synthesize visual information has expanded the horizons of computational problem-solving  \cite {wang2022git} \cite {li2023blip2} \cite {gpt4v} \cite {imp2024}. These breakthroughs are particularly promising for pharmaceutical research and development (R\&D) laboratories, where extracting comprehensive datasets from varied environments is a difficult challenge. The application of state-of-the-art generative AI, with unsupervised and multi-modal capabilities \cite{saacao_segment_2023} \cite{saakirillov2023segany} \cite{saaShilongLiu2023GroundingDM}, has the potential to revolutionize the identification of anomaly in pharmaceutical R\&D labs. \par

In our investigation, we explored the capabilities of two generative AI technologies with promising applications in lab monitoring: multi-modal models and Segment Anything Model (SAM) \cite{kirillov2023segment}. Multi-modal models are capable of understanding images and text to generate relevant textual outputs. SAM, on the other hand, is an innovative image segmentation tool that, given an image and a coordinate, generates a precise mask for the object at that location. Impressively, both models functioned effectively 'out of the box', adapting seamlessly to our unique laboratory environment. While multi-modal models were readily usable to our use case, using SAM required us to formulate a new method by combining traditional computer vision techniques as the prompt was limited to coordinate-based instruction. In our discussion, we contemplate the challenges in achieving a truly automatic lab monitoring system based on this study. We also explore the potential synergy of employing SAM as a vision foundation model (VFM) in concert with multi-modal models.

\section{Material and Methods}
\subsection{lab monitoring System}
The lab monitoring system employed in this study consisted of a quadruped robot equipped with a sophisticated 6-axis arm, which included a gripper and an integrated 4K RGB camera for image data acquisition \cite{SPOTARM}. This high-resolution camera allowed for detailed visual monitoring and data collection within the laboratory environment. The robot and its arm were programmed to navigate and interact with the laboratory environment autonomously.

\begin{figure}[ht]
\centering
    \includegraphics[width=0.49\linewidth]{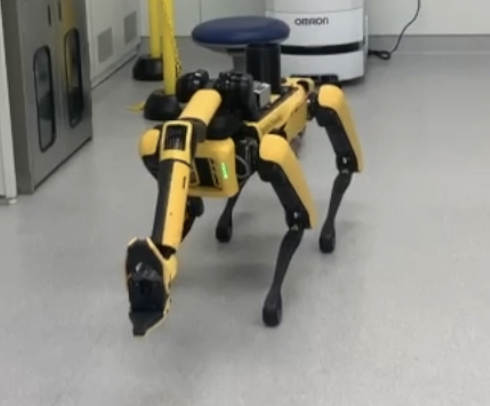}
    \includegraphics[width=0.479\linewidth]{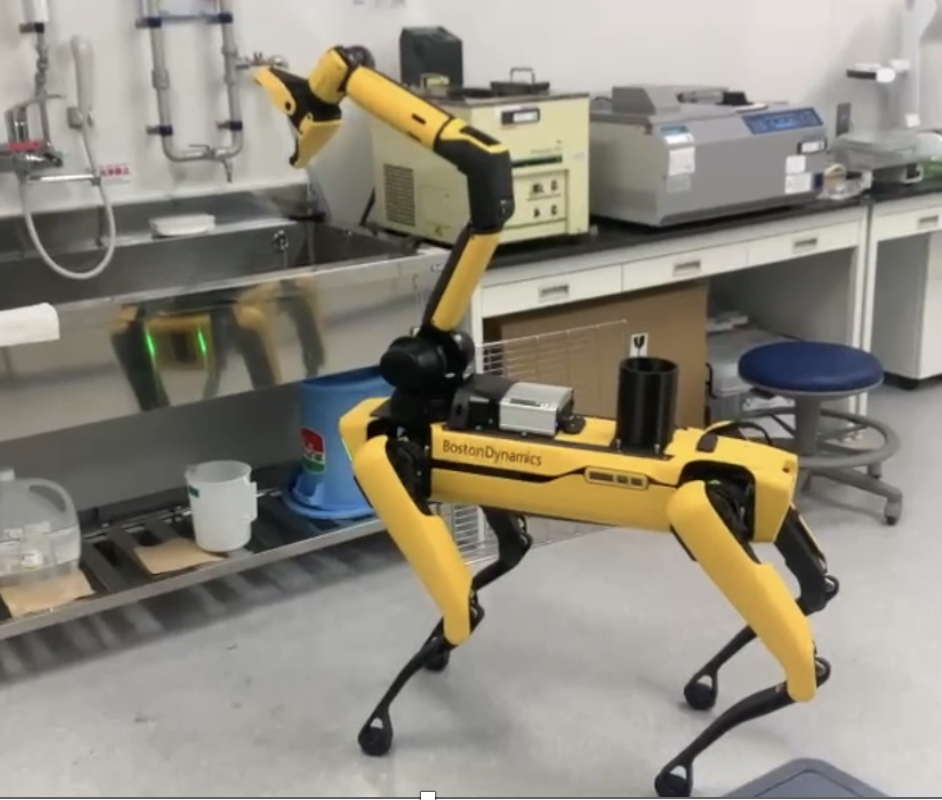}
    \caption{Images of Spot during the lab monitoring process. The ARM enables monitoring of the lab from different angles and heights.}
    \label{fig:lab_monitoring_image} 
\end{figure}

The robot manipulation program was developed with the out-of-box default functionality and Spot SDK provided by Boston Dynamics. This SDK includes an API that facilitates the programming of the control commands for both the robot locomotion and arm manipulation (Fig. \ref{fig:lab_monitoring_image}).

\begin{figure}[ht]
\centering
    \includegraphics[width=0.8\linewidth]{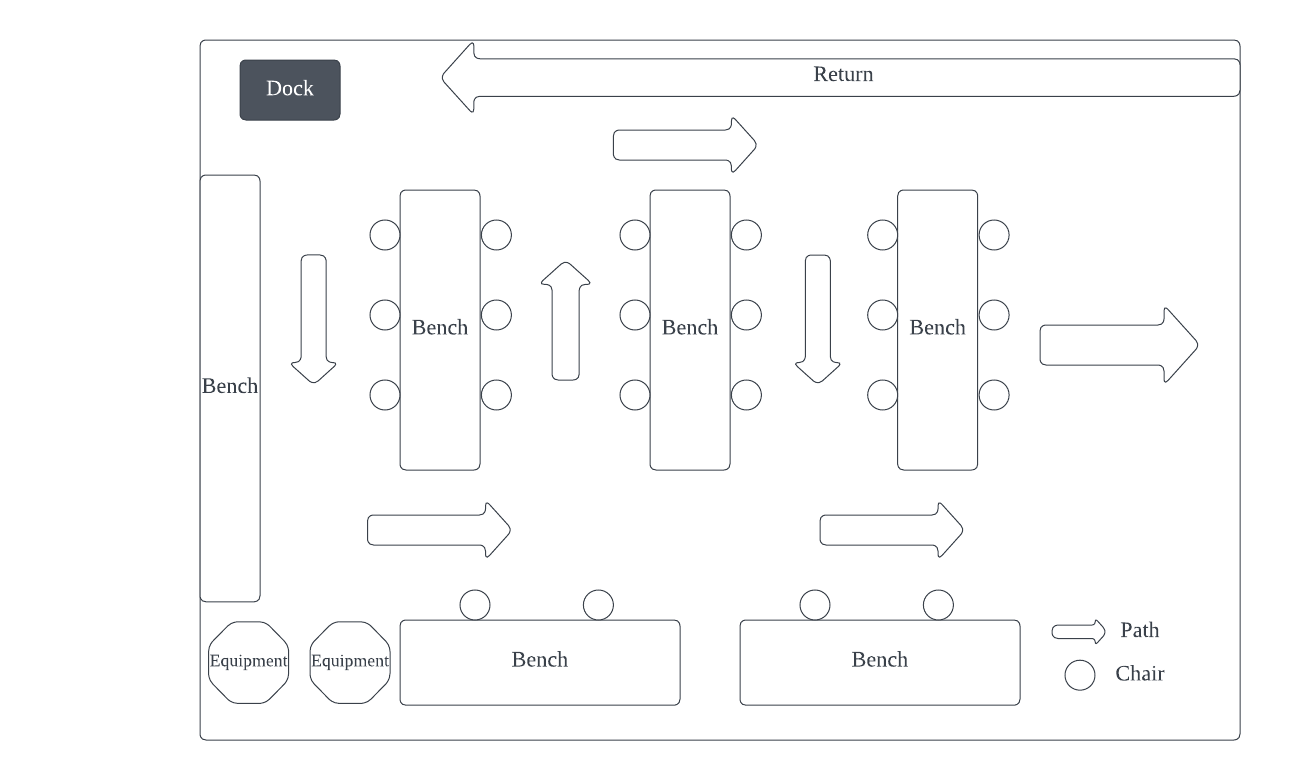}
    \caption{Schematic diagram of the lab monitoring process. The robot follows the path of the arrows taking photographs as it traveled to specified destinations (location of interest)}
    \label{fig:lab_monitoring} 
\end{figure}

To establish the monitoring routine, we first manually guided the robot through the desired route in the laboratory, ensuring it could effectively monitor the locations of interest. During this initial run, we recorded the robot's trajectory and associated actions using the "Autowalk" feature of the Spot SDK. This Autowalk recording was subsequently used to create a repeatable routine that the robot could autonomously execute upon command. A schematic overview of the lab monitoring process is depicted in Fig. \ref{fig:lab_monitoring}

The robot was operated via a dedicated computer that was connected to the same Local Area Network (LAN) as the robot. Upon initiation of the monitoring session by the operator, the Autowalk program was transmitted to and executed by the robot, enabling it to commence its predefined routine. Upon completion of an Autowalk mission, the images captured by the robot's 4K RGB camera were transferred to the computer and processed by the anomaly detection module. 

\subsection{Multi-Modal Model}
The images obtained from the Spot-ARM gripper camera were subjected to processing using the Imp-v1 multimodal small language model (MSLM) \cite {imp2024}. The prompt was adjusted and displayed in the legends of the corresponding figures. The model employed was "MILVLG/imp-v1-3b" sourced from the Hugging Face model repository and the parameters setting were based on it's ModelCard. The torch dtype was set to torch.float16. The tokenizer used in this process was "MILVLG/imp-v1-3b". During the generation process, a maximum of 100 new tokens were allowed. The input image underwent preprocessing using the default method provided by the model.

\begin{figure*}
    \captionsetup[subfigure]{oneside,margin={0.2cm,0.2cm}}
    \centering
    \subfloat[\footnotesize The lab appears to be organized, as the black desk is clean and ready for use.\label{fig:b1}]{
        \includegraphics[width=.24\linewidth]{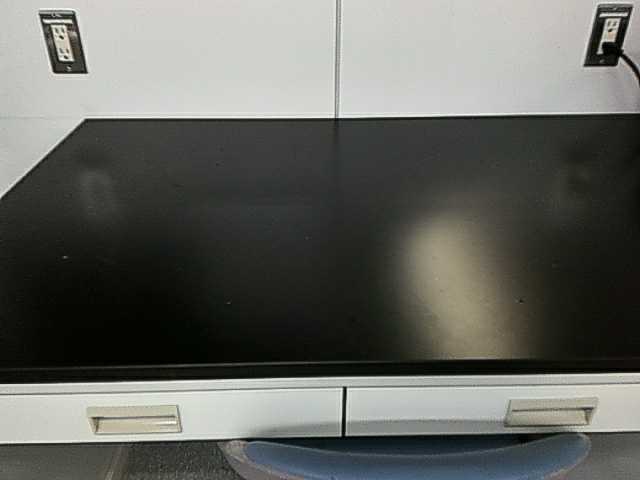}
    }
    \subfloat[\footnotesize The lab appears to be organized, as the black desk is clean and ready for use.\label{fig:b2}]{
        \includegraphics[width=.24\linewidth]{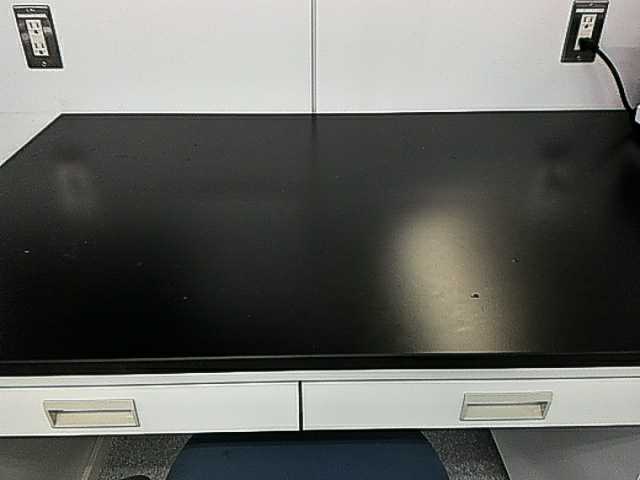}
    }
    \subfloat[\footnotesize The lab appears to be organized, with various items neatly arranged on the table.\label{fig:b3}]{        \includegraphics[width=.24\linewidth]{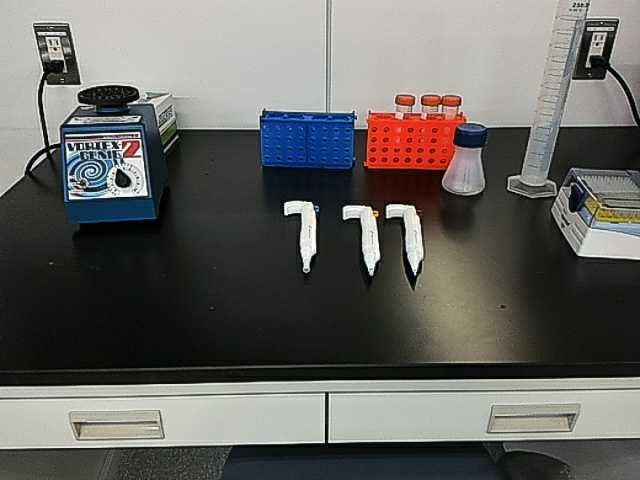}
    }
    \subfloat[\footnotesize The lab appears to be organized, with various items neatly arranged on the table.\label{fig:b4}]{
    \includegraphics[width=.24\linewidth]{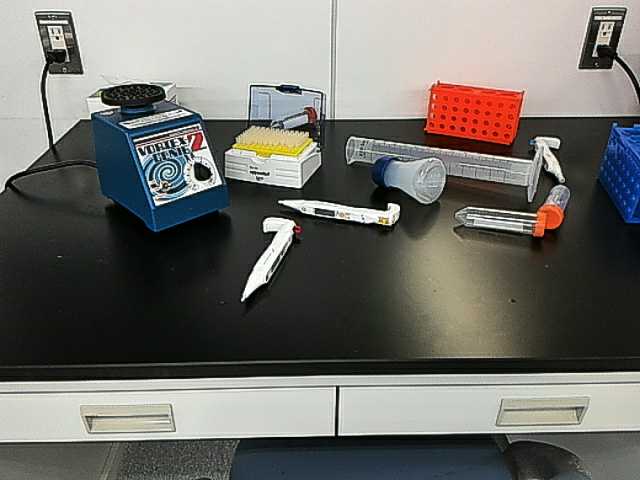}
    }
    \centering
    \hfill
    \subfloat[\footnotesize The lab appears to be organized, with various items neatly arranged on the table.\label{fig:b5}]{
        \includegraphics[width=.24\linewidth]{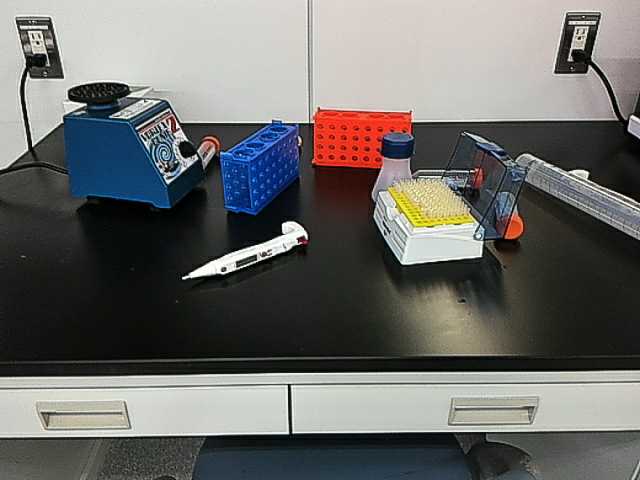}
    }
    \subfloat[\footnotesize The lab appears to be disorganized, with various items scattered on the table, including test tubes, beakers, and other lab equipment.\label{fig:b6}]{
        \includegraphics[width=.24\linewidth]{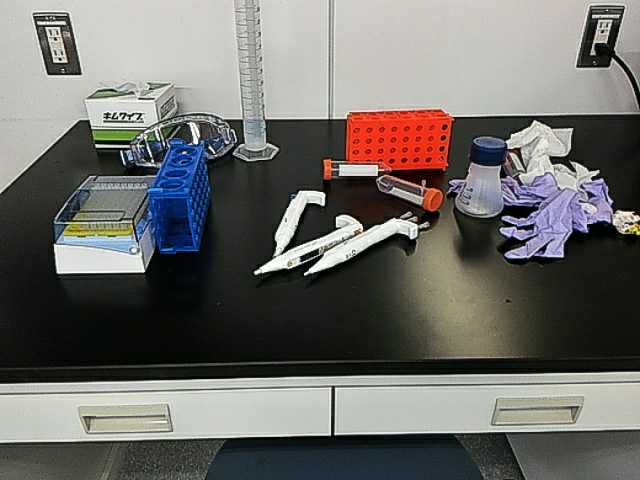}
    }
    \subfloat[\footnotesize The lab appears to be disorganized, with various items scattered on the table, including bottles, test tubes, and other equipment.\label{fig:b7}]{        \includegraphics[width=.24\linewidth]{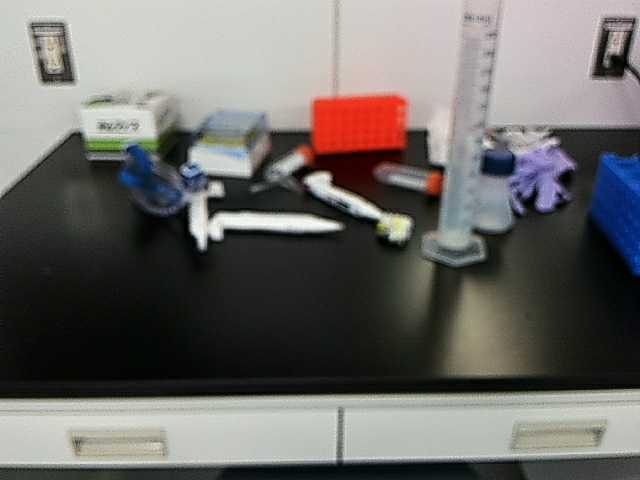}
    }
    \subfloat[\footnotesize The lab appears to be disorganized, with various items scattered on the counter, including a beaker, a test tube, and other lab equipment.\label{fig:b8}]{
    \includegraphics[width=.24\linewidth]{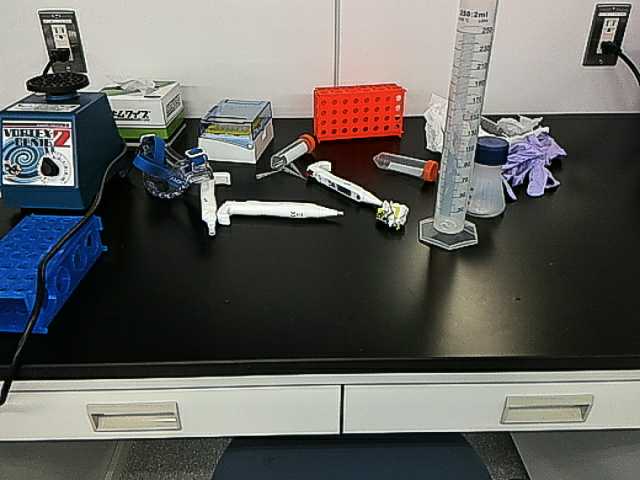}
    }
    \caption{Monitoring of standard laboratory bench. The description of images were generated using the prompt: "A chat between a curious user and an extremely picky inspector for the R\&D lab. The inspector gives detailed answers to the user's questions. USER: <image>\ Is the lab organized or disorganized?: ASSISTANT:".}
    \label{fig:bench}
\end{figure*}

\begin{figure*}
    \captionsetup[subfigure]{oneside,margin={0.2cm,0.2cm}}
    \centering
    \subfloat[\footnotesize The lab appears to be organized, as there is a blue and yellow sign on the wall, which is likely used for safety or identification purposes.\label{fig:p1}]{
        \includegraphics[width=.24\linewidth]{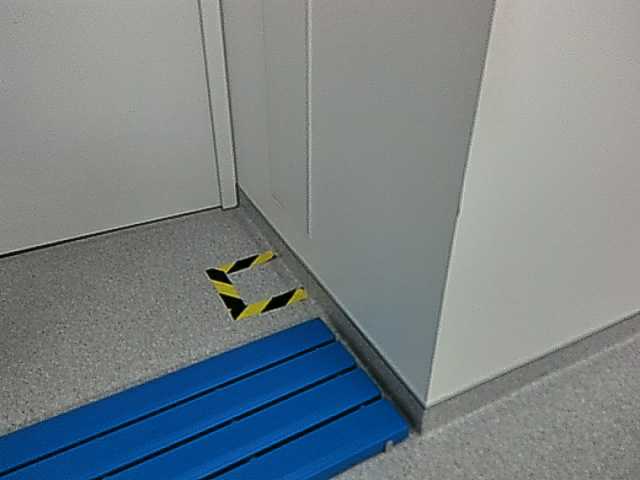}
    }
    \subfloat[\footnotesize The lab appears to be organized, as there is a blue board placed on the floor next to a white wall.\label{fig:p2}]{
        \includegraphics[width=.24\linewidth]{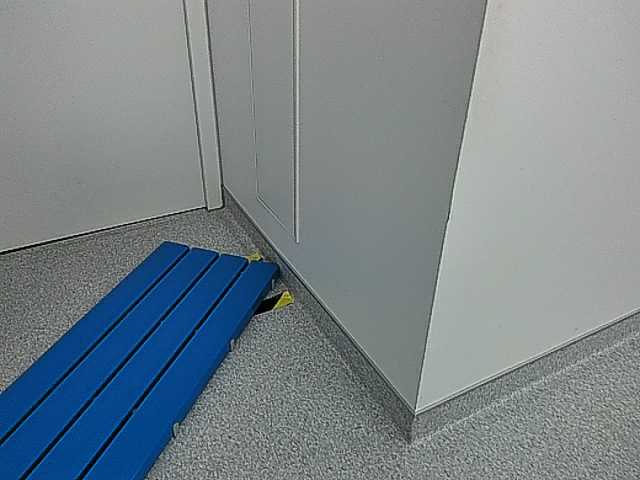}
    }
    \subfloat[\footnotesize The lab appears to be disorganized, as there is a blue board leaning against a wall, which is not a typical arrangement for a lab setting.\label{fig:p3}]{        \includegraphics[width=.24\linewidth]{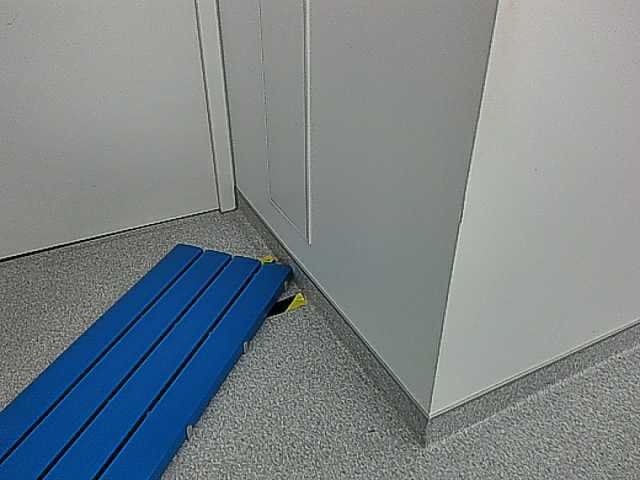}
    }
    \subfloat[\footnotesize The lab appears to be organized, as there is a blue mat with shoes on it, which suggests that the shoes are placed neatly and orderly.\label{fig:p4}]{
    \includegraphics[width=.24\linewidth]{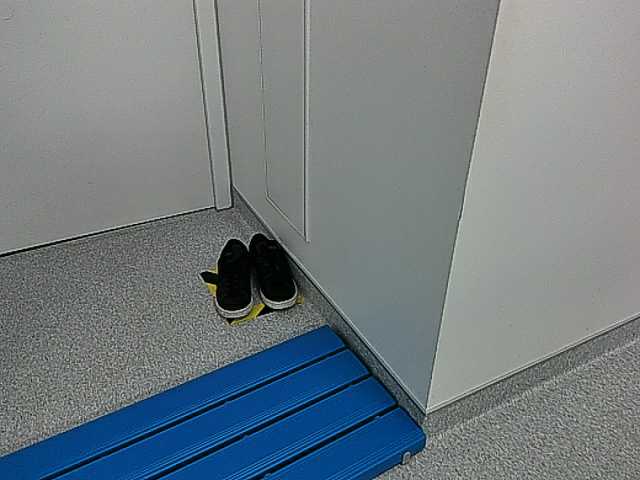}
    }
    \centering
    \hfill
    \subfloat[\footnotesize The lab appears to be organized, as there is a blue and yellow sign on the floor, which indicates a designated area or a specific purpose.\label{fig:p5}]{
        \includegraphics[width=.24\linewidth]{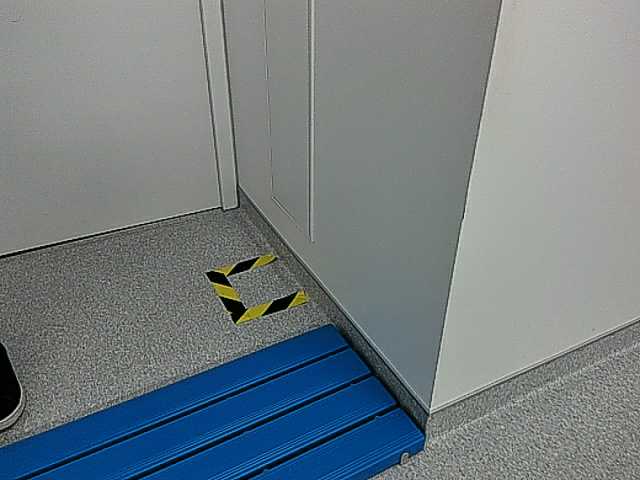}
    }
    \subfloat[\footnotesize \footnotesize The lab appears to be organized, as there is a blue plastic mat on the floor next to a white cabinet, and a bottle of liquid*\label{fig:p6}]{
        \includegraphics[width=.24\linewidth]{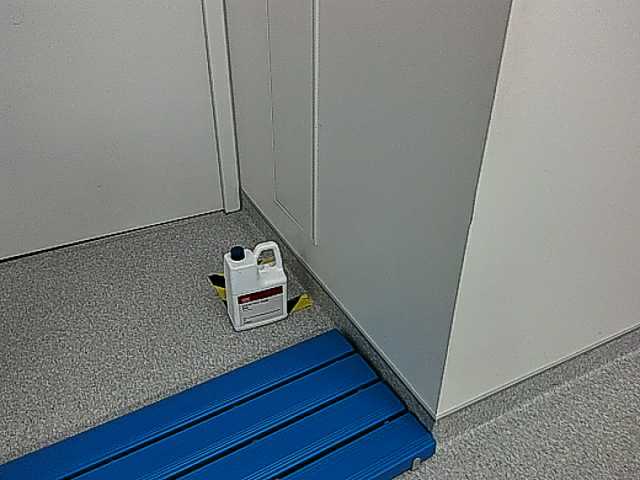}
    }
    \subfloat[\footnotesize The lab appears to be disorganized, as there is a box on the floor next to a wall, and the box is not properly placed or stored.\label{fig:p7}]{        \includegraphics[width=.24\linewidth]{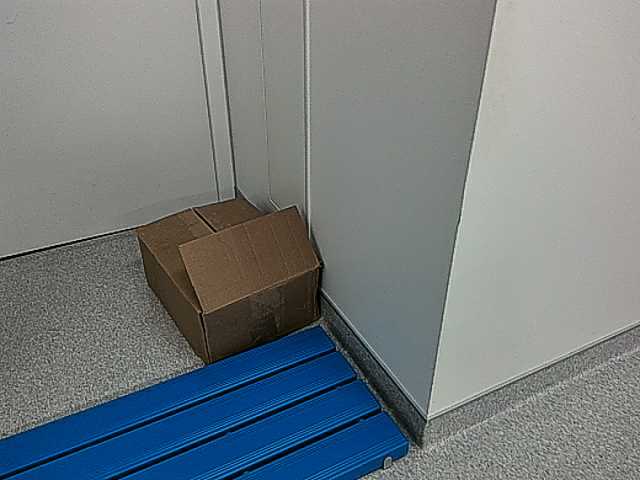}
    }
    \subfloat[\footnotesize The lab appears to be disorganized, as there is a tangled mess of wires on the floor near the doorway.\label{fig:p8}]{
    \includegraphics[width=.24\linewidth]{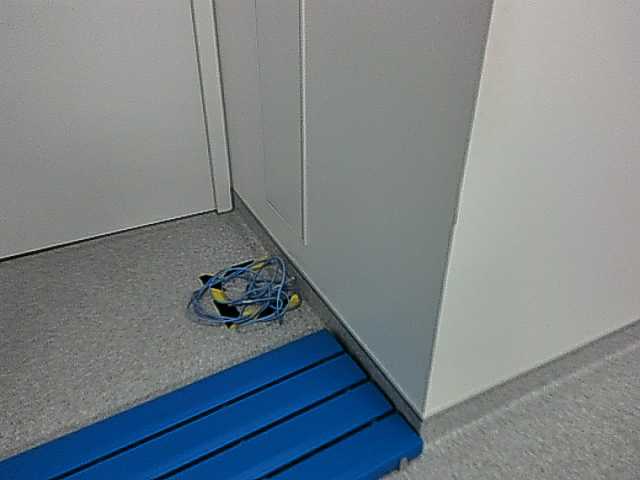}
    }
    \caption{Monitoring of restricted area. The description of images were generated using the prompt: "A chat between a curious user and an extremely picky inspector for the R\&D lab. The inspector gives detailed answers to the user's questions. USER: <image>\ Is the lab organized or disorganized?: ASSISTANT::". *The output was truncated due to its length but is shown as follows: "is placed on the mat. The presence of the mat and the bottle suggests that the lab has designated spaces for storing and handling chemicals, which is a sign of organization.}
    \label{fig:restricted_areas}
\end{figure*}

\subsection{Vision Foundation Model}
The images returned from a monitoring routine were matched with a reference image of the same scene taken with the same routine run at a different time. Anomaly detection was established by looking at inexplicable pixel regions after applying image registration. A new method for the detection of such pixel regions was developed for this end. 

Image registration is the process of mapping two images of a common scene. The process tries to establish correspondence between points or features by transforming one of the images to the other so that the positions of corresponding points or features align. Transformations may encompass translation, rotation, scaling and/or even more intricate deformations. The way transformations are done and how noise is handled depends on the image registration algorithm for which various have been developed. Here we used the optical flow image registration algorithm. The optical flow-based image registration algorithm refines the transformation by adhering to constraints imposed by the optical flow model \cite{lucas1981iterative}. Specifically, it aims to minimize the gray scale net pixel intensity discrepancies between the source and the transformed target images, ensuring consistency with the optical flow model's coherence. A problem arises when an object is only present in one of the images as the correspondence between pixels can no longer be established by it's mapped coordinate given by the algorithm. The method we developed detects such unpairable pixels efficiently, thus enabling the detection of anomalies. The basic idea was that ill matched pixel regions would undergo abnormal transformation or that it would have comparatively high gray scaled net pixel intensity discrepancy with the source image. 
With image registration alone, anomaly detection is limited to the overall gray scale net pixel intensity difference which can be affected by lighting difference and excessive transformation. Also, small differences are hard to detect since it is hard to distinguish if such a small gray scale net pixel intensity difference originated from noise. Therefore, instead of calculating the overall gray scale net pixel intensity difference, we separated the image into regions corresponding to objects using SAM \cite{kirillov2023segment} and calculated the following features to gauge their anomalousness:
\begin{enumerate}
    \item gray scaled net pixel intensity difference between the segmentation region with the corresponding region in the reference image, measured by cosine distance
    \item degree of non-rigid transformation of segmentation region after image registration using Procrustes analysis \cite {gower1975generalized} (disparity)
    \item SAM based signature (segment area) difference between the segmentation and the reference scene
\end{enumerate}

Although the gray scaled net pixel intensity difference feature might seem to be enough for anomaly detection, there are cases where the image registration would minimize this metric by applying transformations such as reducing the size or applying non-rigid transformation of the anomalous object to blend with the surrounding scene. However, since objects in a laboratory environment are mostly solid, non-rigid transformation resulting from the image registration is likely to be an artifact of the optical flow algorithm. In fact, we identified anomalous objects that would have been missed by the gray scaled net pixel intensity difference feature alone, validating its incorporation to our method beyond just theoretical considerations.

While the above mentioned strategy has shown anomaly detection capability to some degree, we have engineered an additional feature aimed to complement and improve the overall performance of the detection system: SAM based signature (segment area) difference between the segmentation and the reference scene. To elucidate the feature in detail, let us first consider its foundational principles. Given a deterministic function that, when provided with the pixel coordinate of an object outputs a value, an object placed identically in both scenes would yield the same values for each of it's pixel coordinate. We can think of the outputs of this function as the signature of the object and we are comparing the signatures. Delving deeper into the specifics, the function we employed operates as follows: we used the SAM algorithm with coordinate of segment as parameter and calculated the segment area size as an output. As any pixel of a given object segments to itself, a single output of the signature function is sufficient to be used as the signature. Thus, the center of each object, which was derived by the object bounding box from SAM was used as the representative of it's respective object and was given as the parameter of the signature function. In other words, we calculated the segment area size of the object and compared the area size when the same coordinate was used in the reference scene. One important detail worth mentioning is that although the obtained scenes are taken from the same position and angle to some extent, the scenes are not aligned perfectly, thus in order to properly compare the signatures, image registration of the scenes was necessary. Finally, using the above three features, we trained an XGBoost classifier to predict anomalous objects.\par
We used python (version $ \geq 3.8 $) with the scipy library (version 1.9.3) for the Procrustes analysis, gray scale net pixel intensity difference quantification and scikit-image library (version 0.21.0) for the optical flow image registration algorithm (registration.optical\_flow\_tvl1 \cite{wedel2009improved}) using the default parameters. In order to compute the transformation by the image registration of each segmentation, the flow fields acquired by the registration algorithm was applied to the segmented objects derived from SAM. The resulting shape of the warped image was analyzed against the original shape using Procrustes analysis to calculate the shape disparity introduced by the image registration. For the gray scale net pixel intensity difference feature we used the cosine method from "scipy.spatial.distance". For image segmentation using SAM (version 1.0), the methods SamPredictor and SamAutomaticMaskGenerator were used. Finally we used the xgboost library (version 2.0.0) for the classifier trained with a learning rate of 0.1, number of estimators to 100, maximum depth of tree to 3, hessian of 1, gamma = 0, subsample ratio of 0.8 and subsample ratio of column when constructing trees to 0.8.

\begin{figure*}[ht]
\captionsetup[subfigure]{oneside,margin={0.2cm,0.2cm}}
\subfloat[\footnotesize There is a white tile floor in the image.\label{fig:subfig1c}]{
    \includegraphics[width=.24\linewidth]{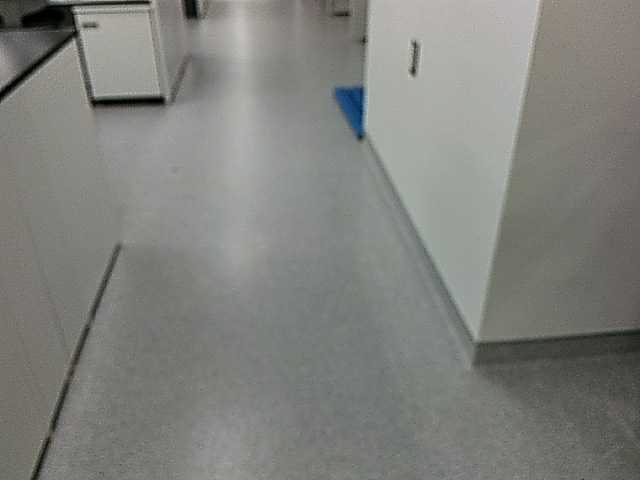}
}
\subfloat[\footnotesize There is a long hallway with a gray floor, and it appears to be empty.\label{fig:subfig2c}]{
    \includegraphics[width=.24\linewidth]{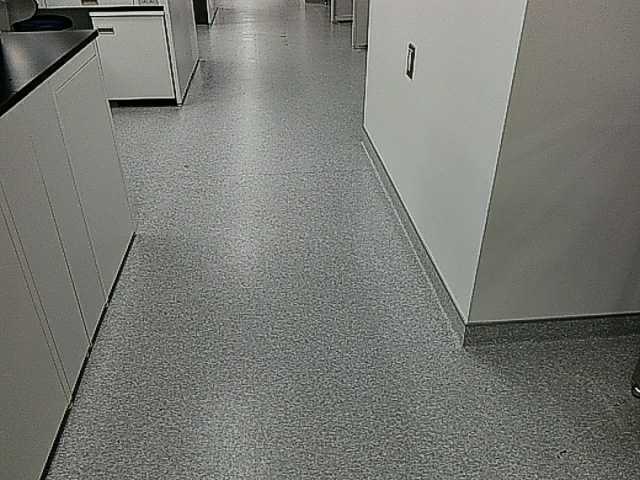}
}
\subfloat[\footnotesize There is a grey tile floor in the image.\label{fig:subfig3c}]{
    \includegraphics[width=.24\linewidth]{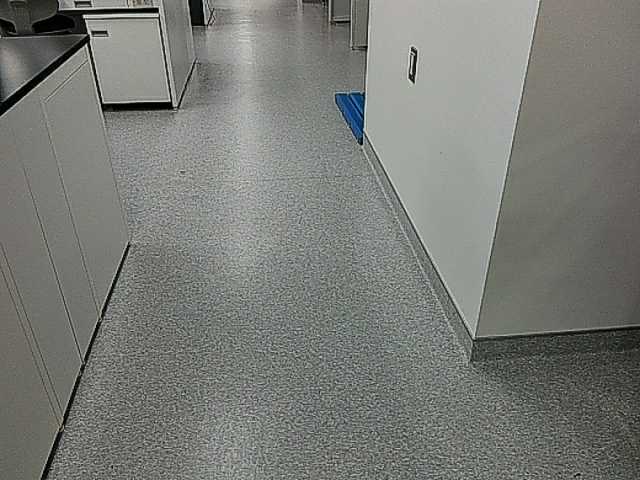}
}
\subfloat[\footnotesize There is a pile of trash on the floor.\label{fig:subfig4c}]{
    \includegraphics[width=.24\linewidth]{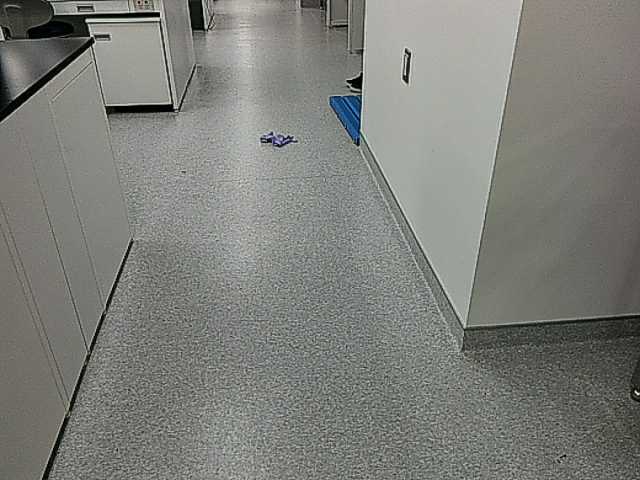}
}
\hfill
\subfloat[\footnotesize There is a pair of shoes on the floor.\label{fig:subfig5c}]{
    \includegraphics[width=.24\linewidth]{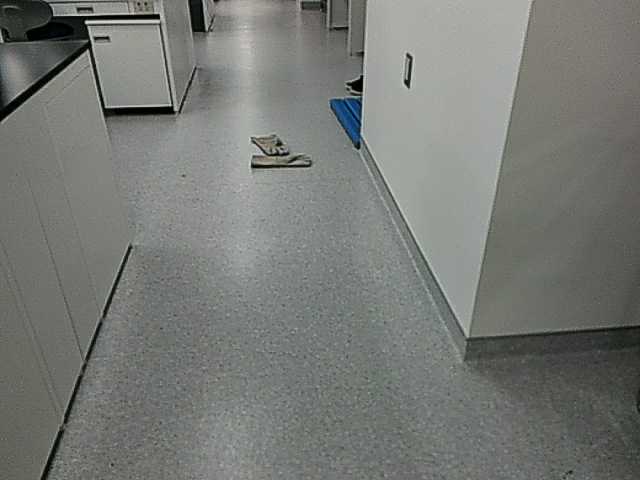}
}
\subfloat[\footnotesize There is a box on the floor.\label{fig:subfig6c}]{
    \includegraphics[width=.24\linewidth]{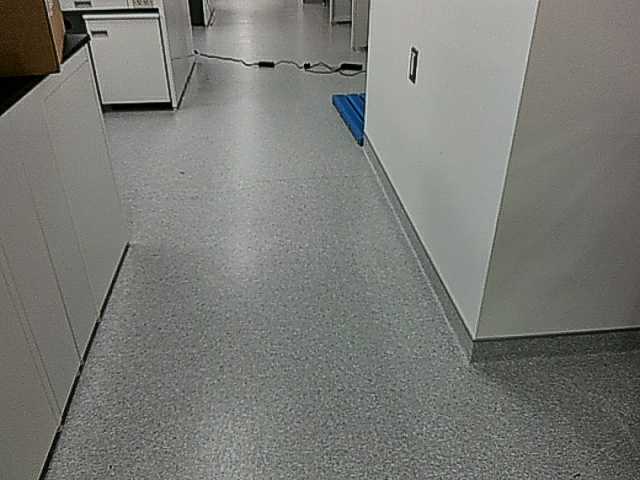}
}
\subfloat[\footnotesize There is a box on the floor.\label{fig:subfig7c}]{
    \includegraphics[width=.24\linewidth]{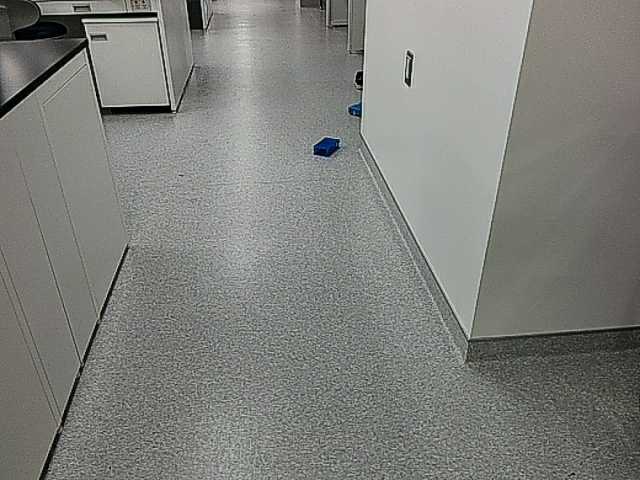}
}
\subfloat[\footnotesize There is a box on the floor.\label{fig:subfig8c}]{
    \includegraphics[width=.24\linewidth]{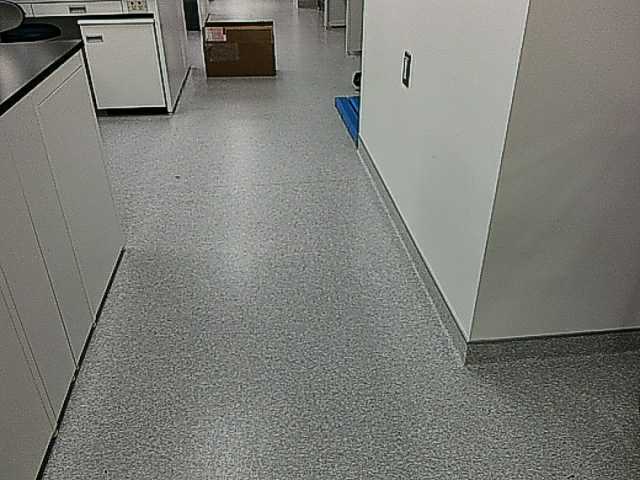}
}
    \caption{Monitoring of laboratory hallway. The description of images were generated using the prompt: "A chat between a curious user and an extremely picky inspector for the R\&D lab. There should be no objects on the floor.The inspector gives detailed answers to the user's questions. USER: <image>\ What is on the floor?: ASSISTANT:"}
    \label{fig:hallway}
\end{figure*}

\begin{figure*}
\captionsetup[subfigure]{oneside,margin={0.2cm,0.2cm}}
\subfloat[\footnotesize There is a stool on wheels in the middle of the room, and a computer is sitting on a desk.]{
    \includegraphics[width=0.24\linewidth]{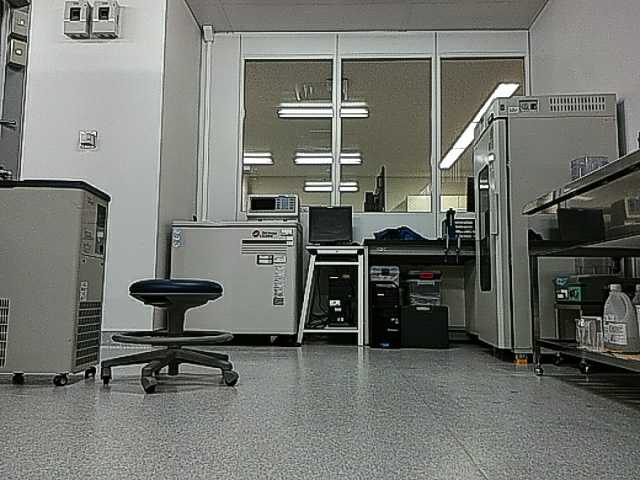}
}
\subfloat[\footnotesize There is a white object on the floor, which appears to be a glove.]{
    \includegraphics[width=.24\linewidth]{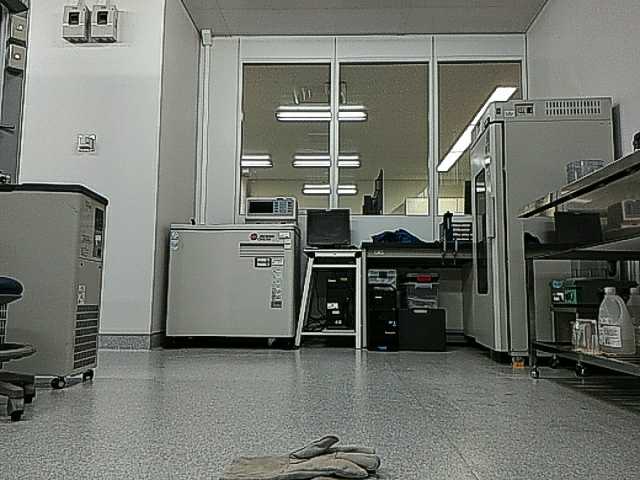}
}
\subfloat[\footnotesize There is a cable on the floor in the room.]{
    \includegraphics[width=.24\linewidth]{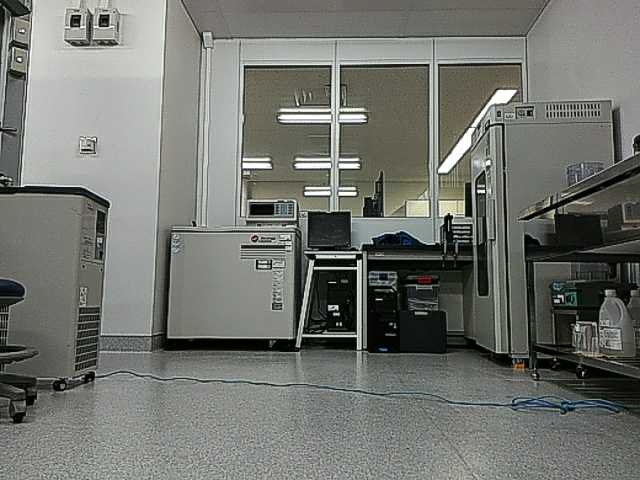}
}
\subfloat[\footnotesize There is a box on the floor in the room.]{
    \includegraphics[width=.24\linewidth]{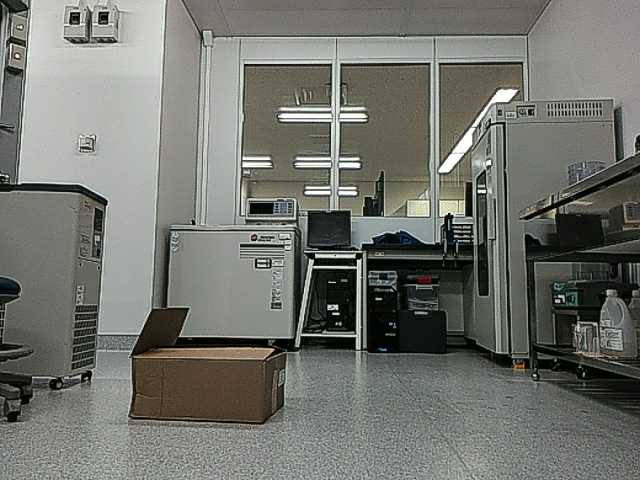}
}
\caption{Monitoring of laboratory floor. The description of images were generated using the prompt: "A chat between a curious user and an extremely picky inspector for the R\&D lab. There should be no objects on the floor.The inspector gives detailed answers to the user's questions. USER: <image>\ What is on the floor?: ASSISTANT:".}
\label{fig:floor}
\end{figure*}

\begin{figure}
    \centering
    \subfloat[\footnotesize Example of SAM output\label{fig:subfigure0}]{
        \includegraphics[width=.48\linewidth]{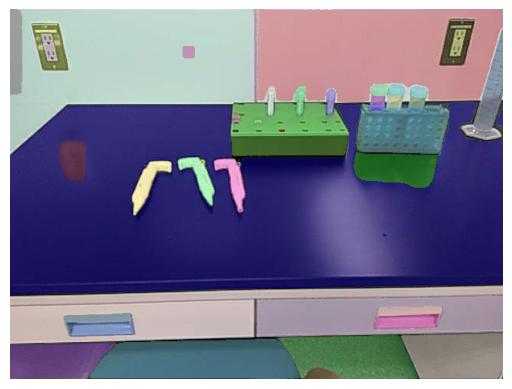}
    }
    \subfloat[\footnotesize Reference Scene\label{fig:subfigure1}]{
        \includegraphics[width=.48\linewidth]{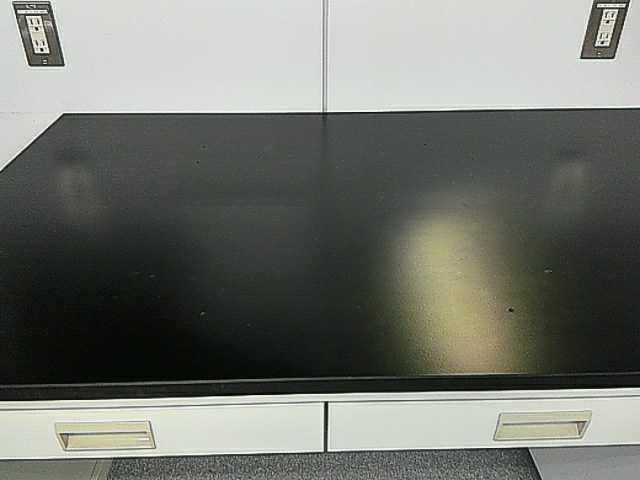}
    }
    \hfill
    \subfloat[\footnotesize Anomaly Label\label{fig:anomaly_obj}]{        \includegraphics[width=.48\linewidth]{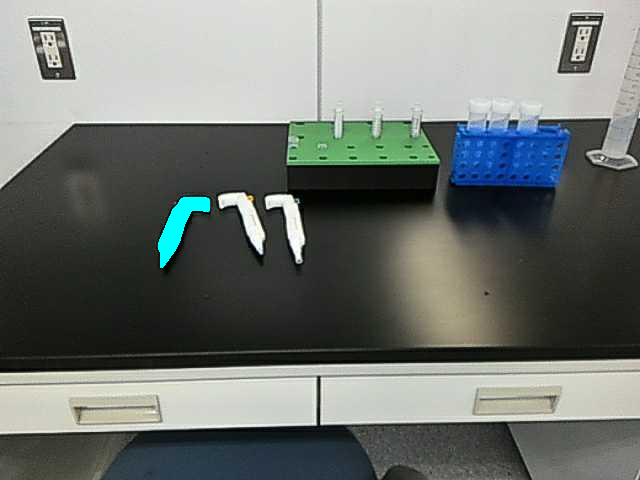}
    }
    \subfloat[\footnotesize Normal Label\label{fig:normal_obj}]{
    \includegraphics[width=.48\linewidth]{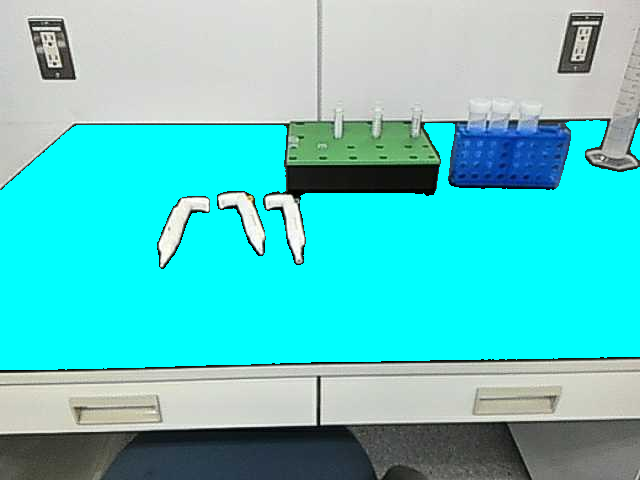}
    }
    \caption{Training data were generated by analyzing the identical scene taken at different time points. One image served as the reference scene, and objects that were not present in the reference were labeled as anomalous. Objects in the non-reference scene were segmented and their segmentation were displayed independently and superimposed on the scene to facilitate efficient analysis. a) Example of segmentation by SAM b) Reference scene c) Segmentation of a pipette superimposed to the scene. This object does not appear in the reference scene and therefore was labeled as anomalous. d) Segmentation of the desk object. Since the desk also appears in the reference scene it was labeled as normal.}
    \label{fig:labeling}
\end{figure}

\section{Result}  \label{sec:Result} 
\subsection{Multi-Modal Model}

To evaluate the feasibility of lab monitoring using a mobile robot and image-to-text model (multi-modal model), various areas with different levels of tidiness and objects were prepared and monitored, including laboratory bench, hallway, floor, and restricted area. The image-to-text model was prompted with the phrase 'A chat between a curious user and an extremely picky inspector for the R\&D lab.' to ensure meaningful generation of output regarding tidiness and anomaly from lab images. This is crucial as industrial laboratory must adhere to stringent organizational and safety standards. We first confirmed that Spot-ARM is able to consistently capture images with high positional reproducibility. This allows for effective monitoring of changes in laboratory environments at specific points of interest (Fig. \ref{fig:bench}a-b).\par
Next, we inquired whether the image-to-text model can perceive the organization of the lab bench. As shown in Fig. \ref{fig:bench}, the model detected the presence or absence of objects on the table and the level of organization. A decision boundary of whether the bench is organized or disorganized needs further refinement, as the model generated the same label for bench with objects that are placed in organized position (Fig. \ref{fig:bench}c) and bench with objects that are placed in disorganized position Fig. \ref{fig:bench}d-e. Nevertheless, for both side of the extremes in terms of the level of organization (Fig. \ref{fig:bench}a-c, f-h), the model was able to generate output that is aligned with human intuition (Fig. \ref{fig:bench}a, b, f-h). For images on restricted area, the model was able to detect presence or absence of objects but inconsistency in output for the level of organization were observed. Specifically, even though the difference between Fig. \ref{fig:restricted_areas}b and c are barely noticeable, the label for the level of organization were organized and disorganized respectively. Interestingly, the image of a liquid bottle with a blue mat was labeled as organized (Fig. \ref{fig:restricted_areas}f). The model stated, "the presence of the mat and the bottle suggests that the lab has designated spaces for storing and handling chemicals, which is a sign of organization". This suggests that the provided context information was insufficient. This also illustrates that a whether a scheme is organized or not is highly depended on the context. The same observation applies to Fig. \ref{fig:restricted_areas}d where the oversight of the yellow and black strip, caused by shoes covering the area, resulted in the removal of critical context information, ultimately leading the model to conclude that this state is in a well-organized state. \par
The model demonstrated the capability to detect the presence or absence of objects on the lab hallway and floor (Fig. \ref{fig:hallway} and Fig. \ref{fig:floor}). However, as observed in previous experiments, there were inconsistencies (Fig. \ref{fig:hallway}a-c) in the output when assessing similar images. 

\subsection{Vision Foundation Model}
In addition to qualitatively analyzing anomalies using an image-to-text model, we assessed whether quantitative information, such as the number of new objects, could be detected using Vision Foundation Models like SAM.
\subsubsection{Data preparation}
A total of 136 objects were identified and segmented utilizing SAM \cite{kirillov2023segment}. Each object was classified as either an anomaly or a normal object based on its presence in the reference scene. Specifically, 60 objects were determined to be anomalous, while the remaining 76 were categorized as normal. Two scenes among the cleanest laboratory workspace scenes where chosen, serving as the baseline for comparison. One scene of each corresponding reference scene was chosen and objects that were absent in their respective reference scene were designated as "anomalies" (Fig. \ref{fig:anomaly_obj}), whereas objects that were consistently present in the reference image were designated as "normal" (Fig. \ref{fig:normal_obj}).

\subsubsection{Two feature analysis}
\begin{figure}
    \centering
    \includegraphics[width=0.8\linewidth]{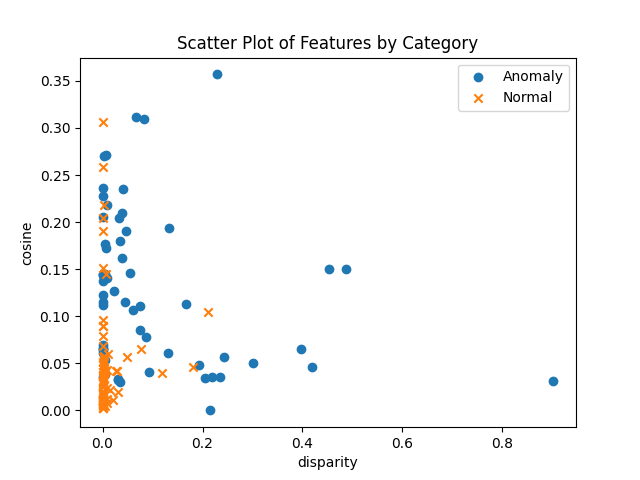}
    \caption{Scatter plot of the non-rigid transformation feature denoted as "disparity" and gray scale net pixel intensity difference feature denoted as "cosine"}
    \label{fig:two_features}
\end{figure}

Initial tests using only the gray scale net pixel intensity difference (cosine) and non-rigid transformation feature (disparity) features revealed a clustering of normal objects towards the bottom-left corner of the feature space whereas anomalous objects were away from the cluster Fig. \ref{fig:two_features}. Despite this pattern, that seemed as though a simple linear model could be used as an effective classifier, a significant number of normal objects were outside the cluster (bottom left), highlighting the complexity of the problem. This challenge led us to develop the third feature, SAM based signature (segment area) difference between the segmentation and the reference scene. 

\begin{figure}
    \centering
    \subfloat[\footnotesize ROC curve of the XGBoost classifier\label{fig:roc_curv}]{
        \includegraphics[width=.80\linewidth]{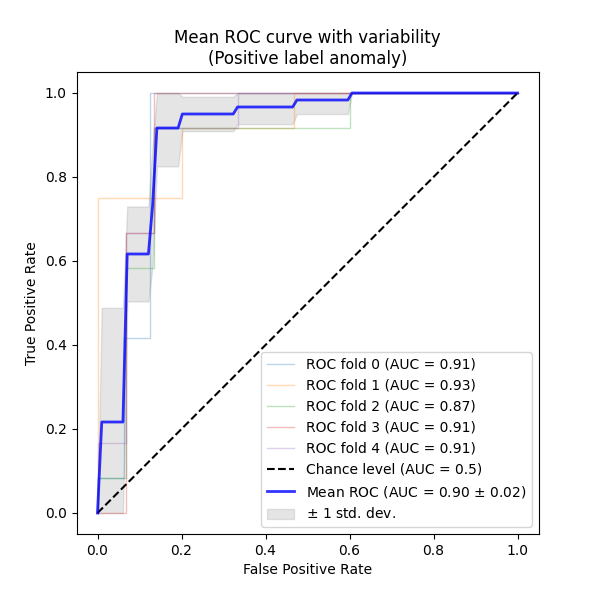}
    }
    \hfill
    \subfloat[\footnotesize Detected anomalous objects shown in red.\label{fig:sample_seg}]{
        \includegraphics[width=.80\linewidth]{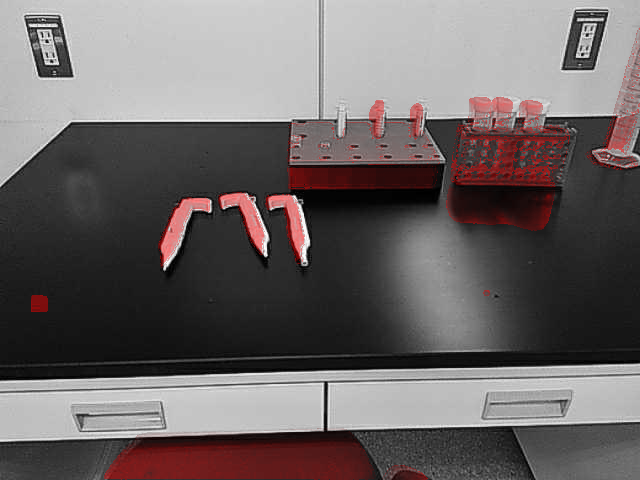}
    }
    \caption{Performance of the classifier was evaluated using 5-fold cross-validation. Sample segmentation results (b) are presented for one of the folds.}
\end{figure}
\subsubsection{Three feature analysis using XGBoost}
To capture the intricate non-linear relation between the three features for classifying normal and anomalous objects, we tested various known methods using five fold cross validation. XGBoost \cite{chen2016xgboost} was the most promising algorithm, given it performed at a mean AUC of $0.9 \pm 0.2$ (Fig. \ref{fig:roc_curv} and \ref{fig:sample_seg}). This result shows that the selected three features are reasonable indicators for detecting new objects, and our approach could potentially be used for anomaly detection in lab monitoring applications.

\section{Conclusion and Discussion}  \label{sec:discussion} 
This study evaluated the feasibility of using a mobile robot and generative AI for lab monitoring. The use of mobile robot for the acquisition of images from routine inspection was shown to be practical despite minimal human intervention. The biggest challenge was the automatic analysis of the images acquired by the robot. We chose to use Generative AI for the potential automatic analysis of the images without training data, which is of extreme convenience and practicality for current and future use cases. Our findings show that multi-modal models were indeed useful at automatic analysis despite having to analyze our lab environment for which it was not specialized at. The multi-modal model successfully detected the presence or absence of objects in various areas, but inconsistencies were observed in assessing the level of organization. Despite this, the model proved useful in identifying inappropriate objects in a laboratory setting. On the other hand, SAM, another AI method that can be readily used without training data, showed high accuracy in segmenting and detecting new objects. Thus, we were able to create a novel method for anomaly detection using SAM as the core component.\par
The ideal automatic analysis should be able to conduct qualitative analysis as demonstrated with the multi-modal models and/or quantitative analysis depending on the laboratory operational rules and guidelines. The remarkable progress in the development of more robust multi-modal models gives hope for the possibility of creating such a model that can consistently and intuitively assess the level of organization both qualitatively and quantitatively. Alternatively, multi-modal models may be better suited to autonomously use traditional computer vision tools, a method demonstrated by \cite {surismenon2023vipergpt}, to accomplish tasks such as accurately and reliably evaluating laboratory conditions, especially now that we have VFM out our disposal.

\section*{Acknowledgements}

This work was funded by Takeda Pharmaceutical Company Limited. We thank our teammates for the fruitful collaboration and their help in reviewing the article: Brian Parkinson, Ádám Wolf, Michael Schwaerzler, Masatoshi Karashima, Seishiro Sawamura, Keiko Yokoyama and Takafumi Oishi.

\section*{Conflict of interest statement}
Shunichi Hato and Nozomi Ogawa are employees of Takeda Pharmaceutical Company Limited, Japan. 

\bibliographystyle{IEEEtran}
\IEEEtriggeratref{25}

\bibliography{references}
\end{document}